\documentclass[10pt,twocolumn,letterpaper]{article}

\usepackage{cvpr}

\usepackage{amsmath,amssymb}

\usepackage{cuted}

\usepackage{currfile}

\usepackage{caption}

\usepackage{booktabs,pifont}

\usepackage{array,booktabs,xcolor,colortbl}
\usepackage[ruled,linesnumbered]{algorithm2e}

\usepackage{orcidlink}
\definecolor{cvprblue}{rgb}{0.21,0.49,0.74}

\def\confName{CVPR}
\def\confYear{2027}

\title{Visual Marking as a Native Reasoning Capability for Multimodal LLMs}

\newcommand{\model}{\textit{SelfMark}}
\usepackage{newfloat}

\DeclareFloatingEnvironment[fileext=lop,name=Prompt,placement=tbp]{prompt}

\author{
    Jiazhen Liu~\orcidlink{0000-0003-0584-4571} \quad
    Yanghao Wang\quad  Chaolei Tan{\quad}
    Long Chen\thanks{Corresponding author: \texttt{longchen@ust.hk}} ~\orcidlink{0000-0001-6148-9709} \\[0.3em]
    The Hong Kong University of Science and Technology \\[0.1em]
}

\begin{document}
\maketitle

\begin{abstract}
Visual reasoning often requires maintaining intermediate spatial states that are difficult to express in text. Existing approaches either rely on external drawing tools or image generators, or encode these states in latent tokens without explicit spatial correspondence to the input image. {The former complicates integration with standard MLLMs, while the latter obscures the spatial structure of intermediate states.} We introduce \model{}, which enables MLLMs to construct and use spatially aligned visual marks directly in their native visual feature space. By treating the visual token grid as a canvas, \model{} establishes explicit spatial correspondence between marks and images. It achieves this through a locate--mark--read procedure, where the model identifies relevant locations, constructs task-dependent mark features, and reads corresponding visual features. To illustrate its generality, we jointly train a single model on spatial planning, dense counting, high-resolution VQA, and line tracing. Surprisingly, \model{} outperforms state-of-the-art MLLM-based methods specialized for individual tasks across six challenging benchmarks. These improvements also persist with a stronger backbone, demonstrating the effectiveness of visual marking across different model capabilities. Overall, \model{} provides a unified approach to visual reasoning by integrating spatially grounded marks directly into the visual feature space of MLLMs.
\end{abstract}
\section{Introduction}
\label{sec:intro}

\begin{figure*}[htbp]
  \centering
  \includegraphics[width=\textwidth]{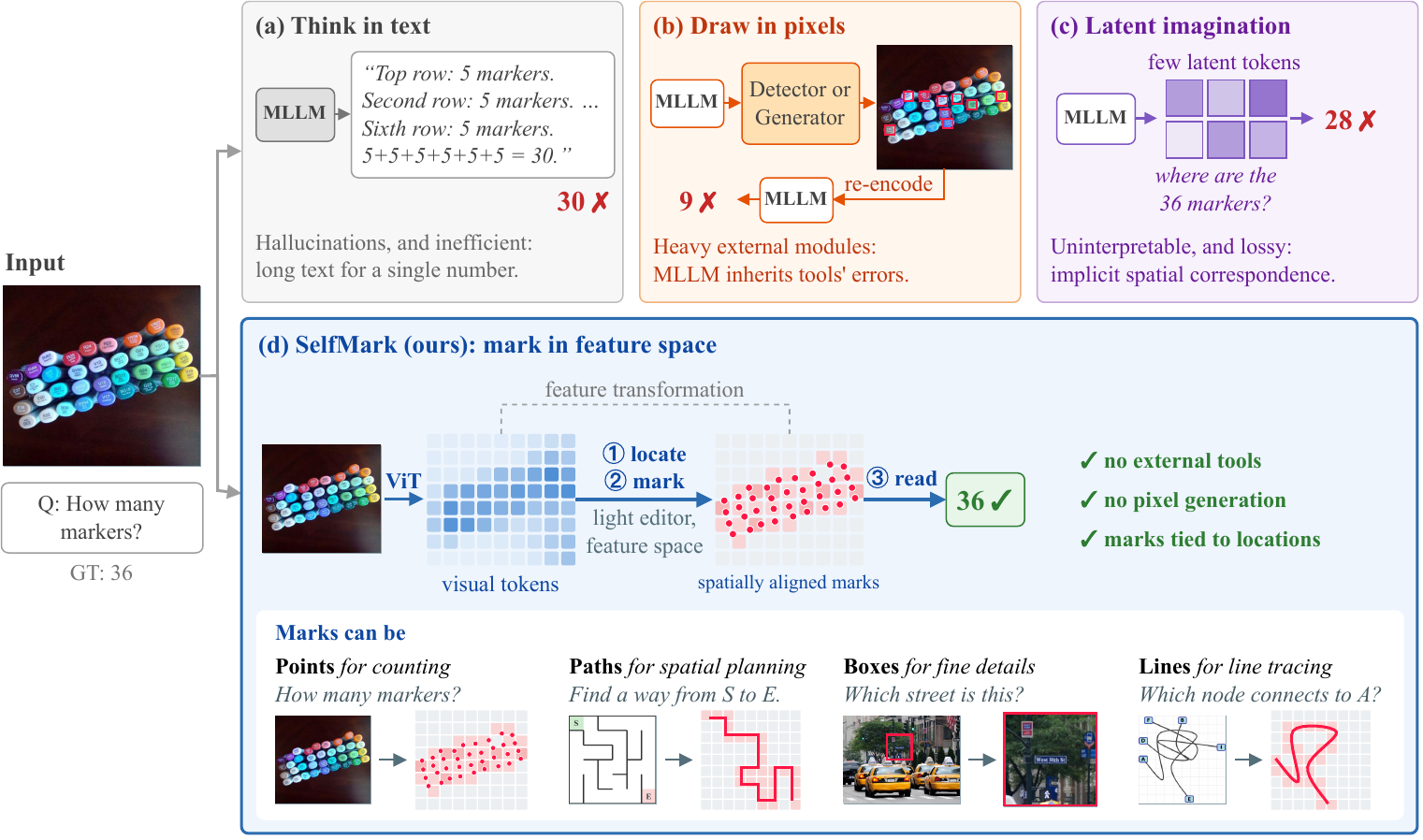}
\caption{
\textbf{Four ways of thinking with an image.}
A simple counting task reveals how different approaches handle
visual evidence.
(a)~\textit{Think in text} translates visual structure into
words, risking plausible but ungrounded descriptions such as
``rows of five.''
(b)~\textit{Draw in pixels} relies on an external vision
module to annotate the image, inheriting its missed detections.
(c)~\textit{Latent imagination} encodes intermediate visual
states in a handful of latent tokens, leaving their spatial
grounding implicit.
(d)~\model{} takes a different route: the MLLM first
\emph{locates} the targets, a lightweight editor \emph{marks}
them in its visual feature space, and the MLLM \emph{reads}
the resulting marks alongside the original visual tokens.
Each dot corresponds to a located object, turning counting
into explicit, inspectable visual reasoning.
The same \emph{locate--mark--read} procedure handles the other tasks with other marks (bottom).
}
  \label{fig:teaser}
\end{figure*}

Multimodal Large Language Models (MLLMs) increasingly solve visual problems through step-by-step reasoning in natural language~\cite{wei2022cot,wang2025vlrethinker}.
However, many visual reasoning tasks involve intermediate spatial states that are hard to represent and update in text alone.
For instance, counting the markers in Fig.~\ref{fig:teaser} requires tracking which markers have been counted and where they are, and solving a maze requires tracking the squares the route has passed through.
For such tasks, reasoning in language alone is prone to errors that accumulate over successive steps.
In contrast, humans often mark the image as they go, dotting each counted object or tracing the route, and then read the marks~\cite{ullman1984visual}.

Inspired by this, recent studies augment textual reasoning with visual intermediate states, which better preserve spatial information than language.
Generally, these approaches fall into two paradigms: 1) Drawing in pixels (\textit{cf.}~Fig.~\ref{fig:teaser}b) makes the state an explicit image that is easy to interpret.
The image is drawn by external tools~\cite{yang2023som,hu2024visualsketchpad,duan2025codeplot,yin2026metareason} or synthesized by unified models~\cite{chern2025thinking,li2025mvot,shi2025mathcanvas,zhou2024transfusion,deng2025bagel}.
{Either way, they rely heavily on additional components or capabilities beyond standard MLLMs}.
In the example of Fig.~\ref{fig:teaser}(b), if the detector does not understand ``markers'' and finds only 9 of the 36, the MLLM inherits this error.
2) Latent imagination (\textit{cf.}~Fig.~\ref{fig:teaser}c) avoids tools and image generation by keeping the state inside a standard MLLM.
The MLLM emits a few latent tokens in its output sequence, as it does with text tokens.
These tokens are trained to match, for example, compressed embeddings of a helper image~\cite{yang2025mirage,xu2026latentgeo} or of query-relevant regions~\cite{li2025latent}, or perceptual and sketch cues~\cite{qin2025covt,zhang2025latentsketchpad}.
Although the MLLM can directly use these tokens for subsequent reasoning, they are highly compressed and lack explicit spatial correspondence with the input image.
As in the example of Fig.~\ref{fig:teaser}(c), there is no token that can tell where the 36 markers are, and their content is hard to inspect.
Recent analyses further find that replacing them with uninformative tokens, or removing them, barely changes the accuracy of several such methods~\cite{li2026imagination,viveiros2026holding,zhu2026silence,guo2026beyond}.
This raises the question of how much visual information they carry.

In summary, both prevalent paradigms fall short in adaptability, reliability, or interpretability. We seek to combine their strengths into a unified visual representation. Specifically, it should reside in the MLLM's feature space with minimal architectural changes, yet maintain explicit spatial correspondence with the input image. These properties would not only support native integration with existing MLLMs and reliable region-level grounding, but also provide clearer training objectives and make the representation's contribution to downstream reasoning easier to verify.

We achieve this with \textbf{\model{}}, a paradigm that places spatial marks directly in the feature space (\textit{cf.}~Fig.~\ref{fig:teaser}d).
The key insight is that the MLLM already represents the input image as a grid of visual tokens, each token corresponding to an image patch.
Placing marks on this grid ensures spatial alignment with the image, much like marking a transparent sheet laid over it.
\model{} constructs and uses these marks in three steps: 
1) In the \emph{locate} step, the MLLM predicts a marking condition for each visual token in a single forward pass~\cite{liu2026STAMP}.
These conditions specify which regions within each patch should be marked.
2) In the \emph{mark} step, a lightweight editor translates these conditions into features of visual marks that the MLLM can interpret.
3) In the \emph{read} step, the MLLM processes the image together with the marks to produce the answer.

Together, these three steps resemble drawing simple annotations on an image and then reading them, except that the marks are constructed directly in the MLLM's feature space.
In \model{}, these annotations take four forms (\textit{cf.}~Fig.~\ref{fig:teaser}d): points for counting, paths for spatial planning, boxes for fine details, and lines for tracing.
All four have simple, well-defined visual forms.
Despite their simplicity, these marks can represent the intermediate spatial states of a wide range of visual reasoning tasks.
Their well-defined appearance also simplifies feature construction, making a lightweight editor sufficient.
At inference time, the MLLM selects the appropriate mark type from the question alone, without manual specification.

To evaluate this flexibility, we instantiate \model{} on Qwen-series MLLMs~\cite{bai2025qwen2} and train a single model for four task families, one per mark type.
Trained only on the data used by prior methods, this model consistently outperforms the same MLLM-based methods.
Ablation studies show that removing the marks at inference time reduces performance to approximately the baseline level, while generating marks from ground-truth conditions further raises accuracy to 99.1\% on spatial planning and 99.8\% on line tracing.
These findings establish spatial marks as meaningful intermediate states that directly improve visual reasoning.

In summary, our main contributions are threefold:
\begin{itemize}
  \item We introduce \model{}, a paradigm that lets a standard MLLM reason with marks written in its own visual feature space and aligned with the image patch by patch.
   \item We realize this paradigm with a locate--mark--read pipeline, in which the MLLM locates and reads while a lightweight editor writes mark features aligned with the visual tokens. It easily adapts to different tasks by changing what is marked.
  \item  We jointly train a single \model{} model on four visual reasoning task families, consistently outperforming the same MLLM fine-tuned on the same data. Ablation studies further validate the effectiveness of the marks.
\end{itemize}

\section{Related Work}
\label{sec:related}

\noindent\textbf{Multimodal Large Language Models.}
MLLMs combine vision encoders with large language models through visual instruction tuning~\cite{alayrac2022flamingo,llava,liu2024improved}.
Models such as Qwen-VL~\cite{bai2023qwenvl,wang2024qwen2,bai2025qwen2,bai2025qwen3vl}, InternVL~\cite{zhu2025internvl3}, and LLaVA-OneVision~\cite{li2024llava} demonstrate strong visual understanding capabilities, including grounding~\cite{peng2023kosmos,chen2023shikra} and high-resolution perception~\cite{wang2024qwen2}.
However, their reasoning remains predominantly textual, making it difficult to track many targets or maintain intermediate spatial states in tasks such as dense counting~\cite{zhang2026wscoc} and spatial planning~\cite{wu2024vsp,yang2025mirage}.

\noindent{\textbf{Visual Reasoning with Intermediate States.}}
Following chain-of-thought reasoning in LLMs~\cite{wei2022cot}, a growing body of work equips MLLMs with intermediate reasoning states in the form of text, pixels, or latent tokens.
Tab.~\ref{tab:taxonomy} organizes representative methods by how these states are produced and whether they correspond to locations in the image.

\underline{\emph{Thinking in Text.}}
This line keeps the entire reasoning process in language; reinforcement learning with verifiable rewards has improved it on math-heavy benchmarks~\cite{wang2025vlrethinker}, and dynamic hybrid training extends it to small MLLMs~\cite{liu2025empowering}.
Spatial information can also be written in text, as object coordinates in Molmo~\cite{deitke2024molmo} or as count ranges narrowed down over a multi-round dialogue in WS-COC~\cite{zhang2026wscoc}.
However, such textual states grow with the number of entities and have no explicit alignment with the image, and the reasoning easily hallucinates structure.

\begin{table}[h]
  \centering
\caption{\textbf{Taxonomy of intermediate states in visual reasoning.}
Columns denote what producing the state relies on: external tools or a unified model that also generates images, or a standard MLLM without image generation.
Rows distinguish intermediate states with explicit correspondence to input-image locations from those represented abstractly.}
  \label{tab:taxonomy}
  \footnotesize
  \setlength{\tabcolsep}{3pt}
  \renewcommand{\arraystretch}{1.25}
  \definecolor{taxhighlight}{RGB}{255,253,220}

  \begin{tabular}{@{}
    >{\raggedright\arraybackslash}m{0.22\linewidth}
    >{\raggedright\arraybackslash}m{0.40\linewidth}
    |
    >{\raggedright\arraybackslash}
      m{\dimexpr0.38\linewidth-4\tabcolsep-\arrayrulewidth\relax}
    @{}}
    \toprule
    \textit{Rely on}
    &
    \textbf{Tools / image generation}
    &
    \textbf{Standard MLLM}
    \\
    \midrule

    \textit{Spatial}\newline
    \textit{correspondence}
    &
    \textit{Tools}: DeepEyes~\cite{zheng2025deepeyes}, Pixel-Reasoner~\cite{su2025pixelreasoner}, Mini-o3~\cite{lai2025minio3}, ZoomEye~\cite{shen2025zoomeye}\newline
    \textit{Unified models}: MVoT~\cite{li2025mvot}, MathCanvas~\cite{shi2025mathcanvas}
    &
    \cellcolor{taxhighlight}
    \textbf{\model{}} \textit{(ours)}:\newline
    grid-aligned marks in visual feature space
    \\

    \cmidrule(l){2-3}

    \textit{Abstract}\newline
    \textit{representation}
    &
    VisProg~\cite{gupta2023visprog}, \newline
    ViperGPT~\cite{suris2023vipergpt}, MM-REACT~\cite{yang2023mmreact}
    &
    \textit{Text}: VL-Rethinker~\cite{wang2025vlrethinker}, WS-COC~\cite{zhang2026wscoc}\newline
    \textit{Latent}: Mirage~\cite{yang2025mirage}, CoLVR~\cite{ding2026colvr}, LatentGeo~\cite{xu2026latentgeo}
    \\

    \bottomrule
  \end{tabular}
\end{table}

\underline{\emph{Drawing in Pixels.}}
This paradigm uses external tools or image generators to construct visual intermediate states, typically fed back to the MLLM as images.
Early tool-based systems compose vision modules or query vision experts for symbolic results~\cite{gupta2023visprog,suris2023vipergpt,yang2023mmreact}.
Later methods overlay numbered marks on segmented regions~\cite{yang2023som} or draw lines, boxes, and auxiliary constructions with code~\cite{hu2024visualsketchpad,duan2025codeplot,yin2026metareason}.
For fine-grained inspection, DeepEyes~\cite{zheng2025deepeyes}, Pixel-Reasoner~\cite{su2025pixelreasoner}, and Mini-o3~\cite{lai2025minio3} learn to crop relevant regions through reinforcement learning, while ZoomEye~\cite{shen2025zoomeye} and attention-based cropping~\cite{zhang2025mllms} do so without training.
Generation-based methods such as MVoT~\cite{li2025mvot}, Thinking with Generated Images~\cite{chern2025thinking}, and MathCanvas~\cite{shi2025mathcanvas} synthesize intermediate images through autoregressive or diffusion-based generation~\cite{zhou2024transfusion,deng2025bagel}, requiring additional generation capabilities and often large-scale retraining.
These images are explicit and usually spatially aligned, but their content remains constrained by how they are produced: external modules may miss targets, cropping only selects regions, and image generation requires capabilities absent from standard MLLMs.

\underline{\emph{Latent Imagination.}}
This paradigm replaces intermediate images with tokens generated by the MLLM itself.
Mirage~\cite{yang2025mirage}, CoLVR~\cite{ding2026colvr}, and LatentGeo~\cite{xu2026latentgeo} supervise continuous tokens with compressed embeddings of helper images or construction diagrams.
Other methods use embeddings of query-relevant regions~\cite{li2025latent} or perceptual cues such as segmentation, depth, and boxes~\cite{qin2025covt,bigverdi2025perception}, or decode latents into sketches~\cite{zhang2025latentsketchpad} or refine them iteratively~\cite{pham2025mcout}.
These methods avoid external tools and image synthesis, but typically generate short token sequences without explicit correspondence to image locations, making their states hard to inspect.
Moreover, replacing or removing these tokens barely affects accuracy in several methods~\cite{li2026imagination,viveiros2026holding,zhu2026silence,guo2026beyond}, consistent with findings that extra tokens can help through computation alone~\cite{pfau2024dots,goyal2024pause}.

\noindent{\textbf{Comparison with \model{}.}}
\model{} occupies the standard, spatially aligned cell of Tab.~\ref{tab:taxonomy}.
Its marks correspond explicitly to image locations, which latent tokens lack, and they are decided by the MLLM itself and written in its own feature space, with no tool or image generator involved.

\section{Method}
\label{sec:method}

\begin{figure*}[t]
  \centering
  \includegraphics[width=\textwidth]{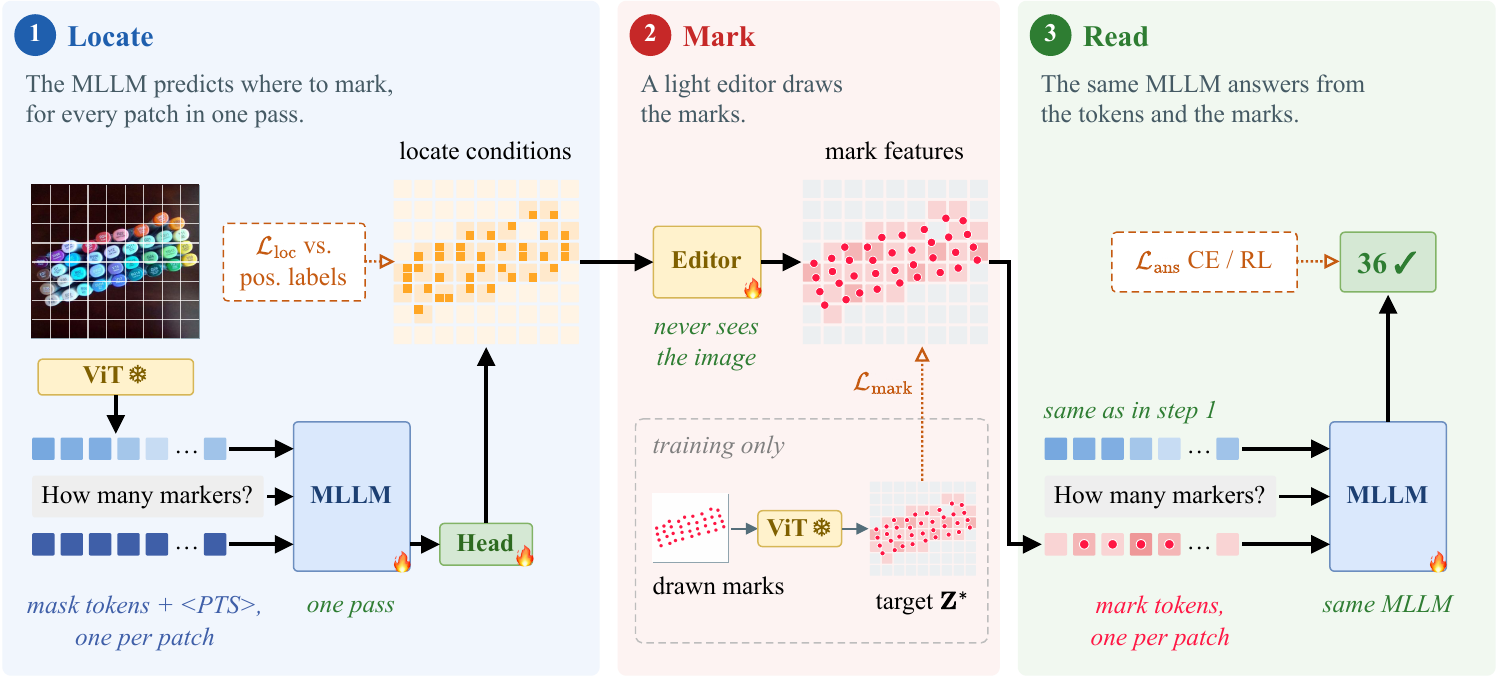}
  \caption{\textbf{Overview of \model{}.}
  \textbf{(1) Locate:} after reading the image and the question, the MLLM emits a type token, here \texttt{<PTS>}, and fills in mask placeholders, one per patch, in a single forward pass; a lightweight head turns them into the locate conditions.
  \textbf{(2) Mark:} a lightweight editor, which never sees the image, translates the conditions into mark features on the same grid as the visual tokens.
  \textbf{(3) Read:} the same MLLM reads the original visual tokens and the question, followed by the mark tokens, and answers.
  Dashed orange parts are used only in training: $\mathcal{L}_{\text{loc}}$ fits the conditions to position labels, $\mathcal{L}_{\text{mark}}$ aligns the mark features with the target ${Z}^{*}$, the frozen-ViT features of the marks drawn on a blank canvas, and the answer is trained with CE and RL; no marks are drawn as pixels at inference.}
  \label{fig:method}
\end{figure*}

In this section, we introduce \model{}, a framework that lets an MLLM reason with marks written in its own visual feature space.
The key challenge is to obtain an intermediate state that is spatially aligned with the image yet produced and read by a standard MLLM.
As depicted in Fig.~\ref{fig:method}, \model{} addresses this challenge with three steps that use the visual tokens of the MLLM as the canvas.
In the \emph{locate} step, the MLLM predicts what should be marked at every patch in a single forward pass.
In the \emph{mark} step, a lightweight editor translates these predictions into mark features on the grid of the visual tokens.
In the \emph{read} step, the same MLLM answers from the original visual tokens together with the marks.
We detail the three steps with counting as the running example in Secs.~\ref{sec:locate}--\ref{sec:read}, and show in Sec.~\ref{sec:training} that the same design and objective carry over to other tasks by changing what is marked.

\noindent\textbf{Notation.}
Given an image $I$ and a question $q$, the vision encoder of the MLLM, \ie, a ViT followed by a projector, maps $I$ to visual tokens $\mathbf{V}=[\mathbf{v}_1,\dots,\mathbf{v}_N]\in\mathbb{R}^{N\times D}$, one for each patch of an $h\times w$ grid with $N=hw$, and $q$ is embedded as text tokens $\mathbf{Q}$.
We write $[\,\cdot\,;\,\cdot\,]$ for concatenation along the sequence.

\subsection{Locate: Predicting Where to Mark}
\label{sec:locate}

The locate step decides, for every patch, what should be marked in it.
Following the all-mask prediction of STAMP~\cite{liu2026STAMP}, \model{} makes this decision for all patches in one forward pass.

\noindent\textbf{Mask Placeholders.}
After the image and the question, we append a block of mask placeholders, one for each patch,
\begin{equation}
  \mathbf{P}=[\mathbf{v}_1+\mathbf{e}_\tau,\dots,\mathbf{v}_N+\mathbf{e}_\tau],
\end{equation}
where $\mathbf{e}_\tau\in\mathbb{R}^{D}$ embeds the type $\tau$ of marks to draw.
Much like the \texttt{<SEG>} token in STAMP~\cite{liu2026STAMP}, the type is decided by the MLLM itself: after reading the question, it emits a type token $\langle\tau\rangle$, \eg, \texttt{<PTS>} for counting, whose embedding serves as $\mathbf{e}_\tau$.
Each placeholder thus carries the content and position of its patch, while $\mathbf{e}_\tau$ tells the MLLM what to locate.
The MLLM processes $\mathbf{S}_{\text{loc}}=[\mathbf{V};\mathbf{Q};\mathbf{P}]$ in a single forward pass, and we take its last-layer hidden states at the positions of the placeholders,
\begin{equation}
  \mathbf{H}=[\mathbf{h}_1,\dots,\mathbf{h}_N]=\mathrm{MLLM}(\mathbf{S}_{\text{loc}})\big|_{\mathbf{P}}\in\mathbb{R}^{N\times D}.
\end{equation}

\noindent\textbf{Locate Conditions.}
A lightweight head $g_\tau$, a two-layer MLP, maps each $\mathbf{h}_i$ to the locate condition of patch $i$,
\begin{equation}
  \mathbf{c}_i=g_\tau(\mathbf{h}_i)\in\mathbb{R}_{\ge 0}^{K},
\end{equation}
which tells where inside the patch marks of type $\tau$ should be drawn.
Specifically, we split each patch into $k\times k$ sub-cells, $K=k^2$, so that the marks can resolve details finer than a patch, and $c_{ij}$ is the occupancy of sub-cell $j$ by the marks, \eg, the expected number of target objects in the sub-cell for counting.
The conditions $\mathbf{C}=[\mathbf{c}_1,\dots,\mathbf{c}_N]$ thus form an occupancy map of the marks on the same grid as $\mathbf{V}$, obtained in one pass.
The head is trained against per-sub-cell labels $y_{ij}$ derived from the annotation,
\begin{equation}
  \mathcal{L}_{\text{loc}}=\frac{1}{N}\sum_{i=1}^{N}\sum_{j=1}^{K}\ell_\tau(c_{ij},y_{ij}),
\end{equation}
where the loss $\ell_\tau$ depends on the type (Tab.~\ref{tab:marks}).

\subsection{Mark: Drawing in Feature Space}
\label{sec:mark}

The locate conditions specify where to mark, yet they are task-specific numbers in a format the MLLM has never read.
For counting, their sum can already estimate the answer; for other tasks, the marked locations must be interpreted together with the image.
Projecting the conditions into the MLLM directly would give features with no constraint to resemble its visual representations.
\model{} therefore converts them into mark features supervised by the vision encoder's representations of rendered marks, and the MLLM reads this visual intermediate state alongside the original image, with no pixels drawn at inference.
Specifically, a lightweight editor $E_\phi$ produces
\begin{equation}
\mathbf{M}=E_\phi(\mathbf{C})\in\mathbb{R}^{N\times D},
\end{equation}
where the mark feature $\mathbf{m}_i$ of patch $i$ lies in the same space as $\mathbf{v}_i$.
$E_\phi$ embeds each condition $\mathbf{c}_i$ with a linear layer, adds a positional embedding of patch $i$, and refines the $N$ embeddings with a few transformer blocks before projecting them to dimension $D$.

The editor takes only the locate conditions as input.
This separates deciding from drawing: the MLLM decides where to mark, and the editor maps these decisions to visual features.
The editor can therefore be trained independently using ground-truth conditions and their corresponding mark-feature targets.

\noindent\textbf{Target of the Marks.}
To keep the marks interpretable and readable, we supervise them with what the MLLM would see if the marks were drawn.
We render the annotated marks, \eg, a red dot at every labeled object, on a blank canvas of the image's size, and encode the canvas with the frozen vision encoder into $\mathbf{Z}^{*}\in\mathbb{R}^{N\times D}$.
The editor is trained to reproduce it,
\begin{equation}
  \mathcal{L}_{\text{mark}}=\frac{1}{N}\big\|\mathbf{M}-\mathbf{Z}^{*}\big\|_F^2.
\end{equation}
Rendering is used only to build this target during training; at inference, the editor writes the marks in feature space directly, and nothing is rendered.

\subsection{Read: Answering with the Marks}
\label{sec:read}

In the read step, the same MLLM answers from the original visual tokens, the question, and the marks,
\begin{equation}
  \mathbf{S}_{\text{read}}=[\mathbf{V};\mathbf{Q};\langle\tau\rangle;\mathbf{M}],
\end{equation}
where the type token $\langle\tau\rangle$ emitted in the locate step introduces the marks.
The original visual tokens are retained, and the marks form a separate map appended to the sequence.
Each mark feature $\mathbf{m}_i$ shares the grid location and feature space of $\mathbf{v}_i$.
For counting, the mark map represents red dots at the predicted object locations, which the MLLM reads together with the image to answer the question.
The answer $y$ is generated autoregressively and trained with the cross-entropy loss $\mathcal{L}_{\text{ans}}=-\log p_\theta(y\mid\mathbf{S}_{\text{read}})$, together with reinforcement learning on a task reward (Sec.~\ref{sec:training}).

\subsection{Training and Extension to Other Tasks}
\label{sec:training}

\noindent\textbf{Mark Types.}
The same three steps extend to different tasks through the mark type $\tau$.
Each type specifies what a locate condition represents, how the annotated marks are drawn to form $\mathbf{Z}^{*}$, and which loss supervises the conditions (Tab.~\ref{tab:marks}).
Points indicate discrete targets, as in counting; paths encode routes, as in spatial planning; boxes select regions whose details matter, as in high-resolution VQA; and lines follow curves already drawn in the image, as in line tracing.
For example, for paths, $c_{ij}$ denotes the fraction of sub-cell $j$ covered by the route and is supervised against the shortest route using binary cross-entropy.
For points, paths, and lines, the editor produces mark features
aligned with the original visual-token grid.
For boxes, we instead crop the located region from the input image, encode it with the MLLM's own vision encoder, and append the resulting visual tokens to the reader, without using the editor or
$\mathcal{L}_{\mathrm{mark}}$.
All types share the MLLM and the read step, and each adds only its embedding $\mathbf{e}_\tau$, prediction head $g_\tau$, and, except for boxes, dedicated channels in the editor's input.

\begin{table}[t]
  \centering
\caption{\textbf{Mark types.} Four mark types used in \model{} and their corresponding training objectives.}
  \label{tab:marks}
  \footnotesize
  \setlength{\tabcolsep}{4pt}
  \begin{tabular}{@{}llll@{}}
    \toprule
    Type & Task & Condition per sub-cell & Loss $\ell_\tau$ \\
    \midrule
    Points & counting & expected objects & Poisson NLL \\
    Paths & spatial planning & route coverage & BCE \\
    Boxes & high-res.\ VQA & region coverage & BCE \\
    Lines & line tracing & curve coverage & BCE \\
    \bottomrule
  \end{tabular}
\end{table}

\noindent\textbf{Which Mark to Draw.}
To let the MLLM decide what to mark without an explicit task label, we introduce one special token per mark type, such as \texttt{<PTS>} for points and \texttt{<PATH>} for paths.
As in STAMP~\cite{liu2026STAMP}, these tokens are added to the vocabulary and generated as part of the MLLM's response.
In all our tasks, the MLLM emits the token right after reading the question, and the emitted token triggers one locate--mark--read cycle, with $\mathbf{e}_\tau$ taken from its embedding.
The tokens are supervised with cross-entropy, and the MLLM learns which type each question calls for.

\noindent\textbf{Objective.}
All tasks are trained in one run with the same objective,
\begin{equation}
  \mathcal{L}=(1-\alpha)\,\mathcal{L}_{\text{ans}}+\alpha\,\mathcal{L}_{\text{RL}}+\beta\,\mathcal{L}_{\text{loc}}+\gamma\,\mathcal{L}_{\text{mark}},
\end{equation}
where $\mathcal{L}_{\text{RL}}$ is GRPO~\cite{shao2024deepseekmath} with a task reward, \eg, whether the plan reaches the goal for spatial planning and a relative-error reward for counting, and $\mathcal{L}_{\text{loc}}$ and $\mathcal{L}_{\text{mark}}$ apply only where mark labels exist.
Following DyME~\cite{liu2025empowering}, we set $\alpha=1$ and apply GRPO when the model samples at least one correct answer. Otherwise, we set $\alpha=0$ and train on the reference answer. This lets each question switch to RL as the model learns to solve it, without a preset switch point.

\section{Experiments}
\label{sec:exp}

\begin{table*}[t]
  \centering
  \newcommand{\sd}[1]{{\tiny$\pm$#1}}
  \newcommand{\grp}[3]{\multicolumn{#1}{@{}l}{\textit{#2}\ (#3)}}
  \caption{\textbf{Comparison with the state of the art.}
  All results of \model{} in (a)--(d) come from a single model, whereas the methods in the middle group of each sub-table are trained for that family alone; the base MLLM of each trained group is given in parentheses.
For \model{}, we report results as mean $\pm$ standard deviation over three seeds. Best results are in bold, and second-best results are underlined.}
  \label{tab:sota}
  \footnotesize
  \begin{minipage}[t]{0.485\textwidth}
    \centering
    \textbf{(a)} Spatial planning: success rate (\%) on VSP\\[3pt]
    \setlength{\tabcolsep}{4.2pt}
    \begin{tabular}{@{}lccccc@{}}
      \toprule
      & \multicolumn{4}{c}{Map size} & \\
      \cmidrule(lr){2-5}
      Method & $3{\times}3$ & $4{\times}4$ & $5{\times}5$ & $6{\times}6$ & Avg. \\
      \midrule
      \multicolumn{6}{@{}l}{\textit{Zero-shot MLLMs}} \\
      Qwen2.5-VL-7B~\cite{bai2025qwen2} & 13.0 & 8.0 & 7.0 & 2.0 & 7.5 \\
      Qwen2.5-VL-32B~\cite{bai2025qwen2} & 28.0 & 23.0 & 14.0 & 11.0 & 19.0 \\
      InternVL3.5-8B~\cite{wang2025internvl35} & 37.0 & 26.0 & 11.0 & 2.0 & 19.0 \\
      LLaVA-OV-7B~\cite{li2024llava} & 6.0 & 6.0 & 3.0 & 0.0 & 3.8 \\
      Qwen3-VL-8B-Thinking~\cite{bai2025qwen3vl} & 46.0 & 30.0 & 32.0 & 23.0 & 32.8 \\
      \midrule
      \grp{6}{Trained for spatial planning}{Qwen2.5-VL-7B} \\
      Mirage~\cite{yang2025mirage} & 90.0 & 84.0 & 67.0 & 55.0 & 74.0 \\
      CoLVR~\cite{ding2026colvr} & {94.0} & \underline{92.0} & \underline{80.0} & \underline{57.0} & \underline{80.7} \\
      \midrule
      \grp{6}{One model for all tasks}{Qwen2.5-VL-7B} \\
      LoRA baseline & \underline{95.0} & 82.0 & 79.0 & 56.0 & 78.0 \\
      \rowcolor{gray!15}
      \model{} & \textbf{99.0} & \textbf{96.0} & \textbf{96.7} & \textbf{80.0} & \textbf{92.9}\sd{0.5} \\
      \bottomrule
    \end{tabular}
  \end{minipage}\hfill
  \begin{minipage}[t]{0.485\textwidth}
    \centering
    \textbf{(b)} Dense counting on FSC-147, whole images without tiling\\[3pt]
    \setlength{\tabcolsep}{2.6pt}
    \begin{tabular}{@{}lcccc@{}}
      \toprule
      & \multicolumn{2}{c}{Val} & \multicolumn{2}{c}{Test} \\
      \cmidrule(lr){2-3} \cmidrule(l){4-5}
      Method & MAE$\downarrow$ & RMSE$\downarrow$ & MAE$\downarrow$ & RMSE$\downarrow$ \\
      \midrule
      \multicolumn{5}{@{}l}{\textit{Zero-shot MLLMs}} \\
      Qwen2.5-VL-7B~\cite{bai2025qwen2} & 41.32 & 114.32 & 38.70 & 146.90 \\
      Qwen2.5-VL-32B~\cite{bai2025qwen2} & 30.16 & 90.84 & 28.84 & 139.39 \\
      InternVL3.5-8B~\cite{wang2025internvl35} & 32.09 & 110.53 & 26.47 & 137.75 \\
      LLaVA-OV-7B~\cite{li2024llava} & 34.47 & 111.04 & 34.59 & 141.42 \\
      Qwen3-VL-8B-Thinking~\cite{bai2025qwen3vl} & 37.40 & 146.17 & 33.27 & 142.65 \\
      \midrule
      \grp{5}{Trained for counting}{LLaVA-OV-7B} \\
      WS-COC-Base~\cite{zhang2026wscoc} & 21.70 & 87.53 & 21.08 & 122.18 \\
      WS-COC~\cite{zhang2026wscoc} & \underline{16.64} & \underline{65.08} & \underline{15.72} & \textbf{105.25} \\
      \midrule
      \grp{5}{One model for all tasks}{Qwen2.5-VL-7B} \\
      LoRA baseline & 21.58 & 86.39 & 17.87 & 127.28 \\
      \rowcolor{gray!15}
      \model{} & \textbf{16.49}\sd{0.29} & \textbf{64.92}\sd{1.90} & \textbf{14.44}\sd{0.40} & \underline{111.18}\sd{3.82} \\
      \bottomrule
    \end{tabular}
  \end{minipage}

  \vspace{8pt}

  \begin{minipage}[t]{0.485\textwidth}
    \centering
    \textbf{(c)} High-resolution VQA: accuracy (\%)\\[3pt]
    \setlength{\tabcolsep}{3.8pt}
    \begin{tabular}{@{}lccc@{}}
      \toprule
      Method & V* & HR-Bench 4K & HR-Bench 8K \\
      \midrule
      \multicolumn{4}{@{}l}{\textit{Zero-shot MLLMs}} \\
      Qwen2.5-VL-7B~\cite{bai2025qwen2} & 75.8 & 69.6 & 65.6 \\
      Qwen2.5-VL-32B~\cite{bai2025qwen2} & 77.0 & 69.8 & 64.8 \\
      InternVL3.5-8B~\cite{wang2025internvl35} & 68.0 & 66.0 & 59.8 \\
      LLaVA-OV-7B~\cite{li2024llava} & 71.3 & 63.5 & 57.0 \\
      Qwen3-VL-8B-Thinking~\cite{bai2025qwen3vl} & 79.8 & \textbf{75.1} & \underline{69.8} \\
      \midrule
      \grp{4}{Trained for high-resolution VQA}{Qwen2.5-VL-7B} \\
    Pixel-Reasoner~\cite{su2025pixelreasoner} & \underline{85.9} & 72.4 & 64.9 \\
      DeepEyes~\cite{zheng2025deepeyes} & 83.0 & \underline{73.3} & 69.7 \\
      \midrule
      \grp{4}{One model for all tasks}{Qwen2.5-VL-7B} \\
      LoRA baseline & 80.8 & 71.8 & 63.7 \\
      \rowcolor{gray!15}
      \model{} & \textbf{86.1}\sd{0.9} & \textbf{75.1}\sd{1.1} & \textbf{71.4}\sd{0.7} \\
      \bottomrule
    \end{tabular}
  \end{minipage}\hfill
  \begin{minipage}[t]{0.485\textwidth}
    \centering
    \textbf{(d)} Line tracing: accuracy (\%) on Tangled Lines, by number of curves\\[3pt]
    \setlength{\tabcolsep}{4.2pt}
    \begin{tabular}{@{}lcccc@{}}
      \toprule
      Method & 3 & 4 & 5 & All \\
      \midrule
      \multicolumn{5}{@{}l}{\textit{Zero-shot MLLMs}} \\
      Qwen2.5-VL-7B~\cite{bai2025qwen2} & 21.7 & 14.5 & 12.4 & 16.0 \\
      Qwen2.5-VL-32B~\cite{bai2025qwen2} & 22.3 & 11.2 & 10.4 & 14.3 \\
      InternVL3.5-8B~\cite{wang2025internvl35} & 19.0 & 15.4 & 10.4 & 14.8 \\
      LLaVA-OV-7B~\cite{li2024llava} & 20.1 & 15.9 & 12.4 & 16.0 \\
      Qwen3-VL-8B-Thinking~\cite{bai2025qwen3vl} & 21.2 & 15.0 & 11.4 & 15.7 \\
      \midrule
      \grp{5}{Trained for line tracing}{Qwen2.5-VL-7B} \\
      KLAL-Base$^\ddagger$ & 20.1 & 11.7 & 9.4 & 13.5 \\
      KLAL$^\ddagger$~\cite{esmaeilkhani2026klal} & \underline{53.3} & \underline{37.5} & \underline{25.7} & \underline{38.4} \\
      \midrule
      \grp{5}{One model for all tasks}{Qwen2.5-VL-7B} \\
      LoRA baseline & 20.9 & 11.7 & 9.9 & 13.9 \\
      \rowcolor{gray!15}
      \model{} & \textbf{64.9} & \textbf{48.0} & \textbf{39.6} & \textbf{50.3}\sd{1.5} \\
      \bottomrule
    \end{tabular}
  \end{minipage}
\end{table*}

\subsection{Common Setup}
\label{sec:setup}

\noindent\textbf{One Model for All Tasks.}
We evaluate four task families, one per mark type (Tab.~\ref{tab:marks}).
Unlike task-specific methods such as CoLVR~\cite{ding2026colvr} and WS-COC~\cite{zhang2026wscoc}, \model{} is trained once on the union of all four training sets.
All its results in Tab.~\ref{tab:sota} come from this single model, which selects the correct mark type from the question alone for every test example. No task label is given at test time.

\noindent\textbf{Compared Methods.}
Each sub-table of Tab.~\ref{tab:sota} has three groups.
The first contains general-purpose MLLMs evaluated in a zero-shot setting, using the same evaluation pipeline as our models.
The second contains the state of the art trained for the family alone (Sec.~\ref{sec:main}).
The third compares \model{} with a mark-free LoRA baseline, each trained jointly on all four task families. The baseline uses the same MLLM, LoRA rank, data, and number of supervised steps, followed by GRPO training.

\noindent\textbf{Implementation Details.}
Unless otherwise stated, we use Qwen2.5-VL-7B~\cite{bai2025qwen2}, the backbone of most baselines, with $896\times896$ inputs ($32\times32$ visual tokens, each with $4\times4$ sub-cells). For locating and reading, we upsample the inputs by $2\times$ when higher spatial resolution is needed.
We fine-tune the language model with separate rank-16 LoRA adapters~\cite{hu2022lora} for locating and reading.
The editor consists of 8 transformer blocks with a hidden dimension of 1,536.
We train on 16 NVIDIA B200 GPUs using AdamW with a batch size of 16.
The learning rates are $10^{-4}$ for the LoRA adapters, $10^{-3}$ for the prediction heads, and $3\times10^{-5}$ for the editor, with $\beta=\gamma=5$.
Each step samples one task in proportion to its training set size.
Additional details are provided in the supplementary material.
\subsection{Main Results}
\label{sec:main}

In each task family of Tab.~\ref{tab:sota}, the single \model{} model outperforms the strongest MLLM-based method trained on the same data and the LoRA baseline.

\noindent\textbf{Spatial Planning with Paths.}
The MLLM plans a route from the start to the goal of a VSP~\cite{wu2024vsp} map around its holes.
We train on the VSP maps of CoLVR, 1,497 after removing duplicates, and test on the 400 maps of Mirage~\cite{yang2025mirage}, 100 for each size from $3\times3$ to $6\times6$.
We report the success rate of the plan.
\model{} succeeds on 92.9\% of the maps, 12.2 points above CoLVR and 14.9 above the LoRA baseline (Tab.~\ref{tab:sota}a).
The margin is largest on the largest maps: on $6\times6$ maps, whose routes are the longest, \model{} reaches 80.0\% against 57.0\% for CoLVR.

\noindent\textbf{Dense Counting with Points.}
The MLLM counts the objects of a given category.
Following WS-COC, we train on the 3,659 training images of FSC-147~\cite{ranjan2021fsc} with their point annotations and report the mean absolute error (MAE) and the root mean squared error (RMSE) on its 1,286 validation and 1,190 test images.
For a fair comparison, every method counts each image as a whole, without test-time tricks such as splitting dense images into tiles.
\model{} reaches the lowest MAE on both splits and the lowest validation RMSE.
On the test split, its MAE of 14.44 is 8\% below the 15.72 of WS-COC and 19\% below the 17.87 of the LoRA baseline.
The gain comes from the marks (Sec.~\ref{sec:analysis}), which help most where counting is hardest.
On the 195 test images with more than 100 objects, \model{} lowers the MAE of the LoRA baseline by a quarter, from 77.7 to 58.0.
WS-COC keeps a lower test RMSE, which a few images with thousands of objects dominate.

\noindent\textbf{High-Resolution VQA with Boxes.}
The MLLM answers questions about small objects in large images.
We follow DeepEyes~\cite{zheng2025deepeyes} and train on the fine-grained questions of its training data.
The base MLLM labels the region of each question by grounding every object that the question mentions, and we keep the 18,644 questions whose region covers less than a quarter of the image.
We test on V*~\cite{wu2024vstar} and HR-Bench~\cite{wang2025hrbench} at 4K and 8K.
The evaluation script of DeepEyes always places the correct answer first among the options, which inflates its accuracy.
For a fair comparison, we use the question files released with the benchmarks, where the correct answer takes varied positions, and evaluate every model in Tab.~\ref{tab:sota}c in the same environment.
\model{} achieves the highest accuracy on all three benchmarks.
It surpasses DeepEyes, which is trained on a superset of our data, by 3.1, 1.8, and 1.7 points, and the LoRA baseline by 5.3, 3.3, and 7.7 points.
A single zoom into the located region is enough, since the marked region contains the queried object on most V* questions.
The gain over the LoRA baseline is largest on HR-Bench 8K, whose images lose the most detail at the input resolution.

\noindent\textbf{Line Tracing with Lines.}
The MLLM follows a curve through a tangle of curves to the box at its other end.
The Line Tracing set of KLAL~\cite{esmaeilkhani2026klal} is saturated, since a LoRA fine-tune exceeds 96\% on it and 206 of its 210 test images are redrawn training images (supplementary material).
We therefore build Tangled Lines, where each $768\times768$ image holds 3 to 5 smooth curves that join lettered boxes on its border, and the question asks which box is joined to a given one.
We generate 30,000 training and 600 test images from disjoint random seeds, with a chance accuracy of about 15\%, and provide the test set and the generator in the supplementary material.
We re-implement KLAL on the same data (KLAL$^\ddagger$), which trains a LoRA of the same rank on line tracing alone with KLAL loss, and KLAL-Base$^\ddagger$ drops this loss.
\model{} reaches 50.3\%, 11.9 points above KLAL$^\ddagger$, and the lead grows to 13.9 points on images with five curves (Tab.~\ref{tab:sota}d).
With the Qwen2.5 backbone, KLAL-Base$^\ddagger$ and the LoRA baseline both remain at chance after answer-only fine-tuning, below the zero-shot base MLLM. Answer supervision supplies only the endpoint label, without indicating the path leading to it. \model{}'s line marks instead highlight the target curve, allowing the MLLM to identify its endpoint directly.

\begin{table}[t]
  \centering
  \caption{\textbf{Ablations on base MLLMs.} \model{} with different base MLLMs against the LoRA baseline. }
  \label{tab:base}
  \footnotesize
  \setlength{\tabcolsep}{3.2pt}
  \begin{tabular}{@{}lcccccc@{}}
    \toprule
    & VSP & FSC-147 & V* & \multicolumn{2}{c}{HR-Bench} & Tangled \\
    \cmidrule(lr){5-6}
    & Acc. & MAE$\downarrow$ & Acc. & 4K & 8K & Acc. \\
    \midrule
    \multicolumn{7}{@{}l}{\textit{Qwen2.5-VL-7B}} \\
    LoRA baseline & 78.0 & 17.87 & 80.8 & 71.8 & 63.7 & 13.9 \\
    \rowcolor{gray!15}
    \model{} & \textbf{92.9} & \textbf{14.44} & \textbf{86.1} & \textbf{75.1} & \textbf{71.4} & \textbf{50.3} \\
    \multicolumn{7}{@{}l}{\textit{Qwen3.5-9B}} \\
    LoRA baseline & 90.5 & 14.12 & 86.1 & 73.8 & 67.1 & 71.8 \\
    \rowcolor{gray!15}
    \model{} & \textbf{98.8} & \textbf{11.50} & \textbf{88.6} & \textbf{79.7} & \textbf{76.4} & \textbf{82.2} \\
    \bottomrule
  \end{tabular}
\end{table}
\begin{table}[t]
  \centering
\caption{\textbf{Effect of visual marks on task performance.}
The upper rows compare inference-time interventions, while the bottom rows evaluate separately trained variants. Dashes indicate unavailable or inapplicable settings.}
  \label{tab:marks_read}
  \footnotesize
  \setlength{\tabcolsep}{3.4pt}
  \begin{tabular}{@{}lcccccc@{}}
    \toprule
    & VSP & FSC-147 & V* & \multicolumn{2}{c}{HR-Bench} & Tangled \\
    \cmidrule(lr){5-6}
    Marks & Acc. & MAE$\downarrow$ & Acc. & 4K & 8K & Acc. \\
    \midrule
    \rowcolor{gray!15}
    Predicted & 92.9 & 14.44 & 86.1 & 75.1 & 71.4 & 50.3 \\
    None & 76.8 & 16.74 & 80.3 & 68.9 & 64.8 & 13.0 \\
    Mirrored&	76.3&	14.41&	79.4&	73.0&	64.3&	17.4 \\
{\color{gray!60} Ground Truth} &
{\color{gray!60} 99.1} &
{\color{gray!60} 6.35} &
{\color{gray!60} 93.9} &
{\color{gray!60} --} &
{\color{gray!60} --} &
{\color{gray!60} 99.8} \\
    \midrule
    w/o marking & 72.6 & 15.00 & 79.3 & 71.1 & 63.0 & 46.0 \\
    w/o $\mathcal{L}_{\text{mark}}$ & 87.3 & 15.31 & -- & -- & -- & 15.0 \\
     \color{gray!60}\quad + Ground Truth & \color{gray!60}87.5 & \color{gray!60}15.32 & \color{gray!60}-- & \color{gray!60}-- & \color{gray!60}-- & \color{gray!60}13.8 \\

    \bottomrule
  \end{tabular}
\end{table}

\begin{figure*}[t]
  \centering
  \includegraphics[width=0.8\textwidth]{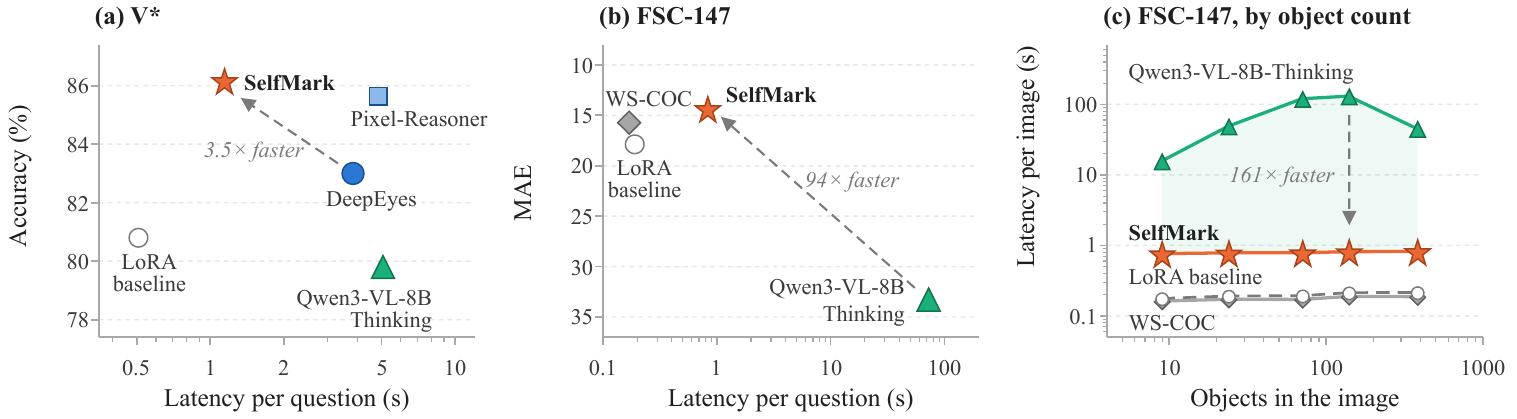}
\caption{\textbf{Inference cost.} (a, b) Performance \vs latency on V* and FSC-147 (upper left is better). (c) Latency \vs object count.}
  \label{fig:eff}
\end{figure*}

\noindent\textbf{Another Base MLLM.}
We further evaluate \model{} and the LoRA baseline on Qwen3.5-9B~\cite{qwen2026qwen35} using the same data and training schedule.
Images are resized to $1024\times1024$ to preserve the $32\times32$ token grid, and the editor is trained from scratch.
\model{} outperforms the LoRA baseline across all six benchmarks (Tab.~\ref{tab:base}).
Notably, the stronger backbone raises the line tracing baseline from 13.9\% to 71.8\%, yet \model{} further improves it to 82.2\%.
These results show that the gains from \model{} carry over to a stronger backbone and persist even when answer-only supervision is already effective.

\subsection{Analysis}
\label{sec:analysis}

\noindent\textbf{Do the Answers Depend on the Marks?} We remove the predicted marks at test time by giving the editor empty conditions or, for box marks, a blank crop (Tab.~\ref{tab:marks_read}, second row). Performance declines on every benchmark. Accuracy falls by 16.1 points on spatial planning and 37.3 points on line tracing, and counting MAE rises from 14.44 to 16.74. Drawing the same marks at their mirror positions, or zooming into the mirrored region for boxes (third row), lowers spatial planning, line tracing, V*, and HR-Bench 8K to about the level without marks and leaves counting unchanged. HR-Bench 4K drops less, as most of its located regions lie near the vertical midline. In related interventions on several latent-imagination methods, replacing the latent tokens with uninformative ones or removing them barely changes accuracy~\cite{li2026imagination,viveiros2026holding,zhu2026silence,guo2026beyond}. Our ablation shows that reading the predicted marks does help \model{} answer the questions. We also complement these quantitative results with visualizations in the supplementary material.

\noindent\textbf{Marks Provide Benefits Beyond Dense Supervision.}
To test whether the gains come from dense localization supervision alone, we train a variant that applies the same supervision to the reader without providing marks (Tab.~\ref{tab:marks_read}, fifth row).
Compared with \model{}, its accuracy drops by 20.3 points on spatial planning and 4.3 points on line tracing, with consistent declines on the other tasks.
We further train the editor from scratch without $\mathcal{L}_{\text{mark}}$, allowing it to learn mark features solely from answer supervision.
This reduces spatial planning accuracy from 92.9\% to 87.3\% and line tracing accuracy from 50.3\% to 15.0\%, while counting MAE rises from 14.44 to 15.31.
Moreover, replacing them with marks from ground-truth conditions (sixth row $\to$ last row) leaves its performance unchanged, indicating that the reader does not use them.
These results show that the marks provide benefits beyond dense supervision, with visual feature alignment playing a particularly important role in line tracing.

\noindent\textbf{Where Do the Errors Come From?}
We assess the potential gains from better localization by generating marks from ground-truth conditions (Tab.~\ref{tab:marks_read}).
Performance improves on all four tasks, with spatial planning and line tracing both exceeding 99\% accuracy.
For counting, reading the predicted marks gives an MAE of 14.44.
Using ground-truth conditions lowers it further to 6.35, suggesting substantial room for further gains through a more accurate locate step.

\noindent\textbf{What Do the Marks Cost?}
We measure end-to-end inference latency on one B200 GPU
at batch size one (Fig.~\ref{fig:eff}). Localization and marking add
about 0.6 s over direct LoRA answering on V* and FSC-147.
Nevertheless, \model{} is 3.5--4.5$\times$ faster than
DeepEyes and Pixel-Reasoner on V*, and 94$\times$ faster
than Qwen3-VL-8B-Thinking on FSC-147. These gains come
from parallel localization and fewer autoregressive
decoding steps, with counting latency remaining nearly
constant across object counts.

\section{Conclusion}
We have introduced \model{}, a paradigm that represents visual intermediate states as spatial marks in an MLLM's own feature space. Its locate--mark--read procedure combines explicit spatial correspondence with visual features that the MLLM can read directly. At inference time, the model locates all marks in a single forward pass and constructs them without external tools or pixel generation. A single jointly trained model handles four task families and outperforms matched baselines across six benchmarks. The gains are often larger on more challenging cases and persist with a stronger backbone. Removing the marks consistently degrades performance, confirming their importance in guiding the model's reasoning. Experiments with ground-truth conditions reveal substantial room for further improvement, particularly through more accurate localization. Together with the reduced autoregressive decoding cost, these findings support \model{} as an effective, inspectable, and efficient representation for multimodal reasoning.

\definecolor{cvprblue}{rgb}{0.21,0.49,0.74}

\definecolor{pmframe}{RGB}{207,216,220}
\definecolor{pmback}{RGB}{250,251,252}
\definecolor{pmuser}{RGB}{31,95,174}
\definecolor{pmuserbg}{RGB}{227,236,247}
\definecolor{pmours}{RGB}{46,125,50}
\definecolor{pmoursbg}{RGB}{226,240,227}
\definecolor{pmbase}{RGB}{84,110,122}
\definecolor{pmbasebg}{RGB}{236,239,241}
\definecolor{pmtoken}{RGB}{198,40,40}
\definecolor{pmmarks}{RGB}{230,100,0}
\newcommand{\pmchip}[3]{{\setlength{\fboxsep}{1.2pt}\colorbox{#1}{\strut\textcolor{#2}{\textbf{#3}}}}}
\newcommand{\pmuser}{\pmchip{pmuserbg}{pmuser}{User}}
\newcommand{\pmours}{\pmchip{pmoursbg}{pmours}{SelfMark}}
\newcommand{\pmbase}{\pmchip{pmbasebg}{pmbase}{LoRA}}
\newcommand{\pmtok}[1]{\textcolor{pmtoken}{\texttt{\textbf{#1}}}}
\newcommand{\pmvis}[1]{\textcolor{pmmarks}{\textit{[#1]}}}
\newcommand{\pmcut}[1]{\textcolor{gray}{\textit{[#1]}}}
\newcommand{\pmbox}[1]{\texttt{\textbackslash boxed\{#1\}}}
\newcommand{\pmtask}[1]{\par\smallskip\noindent\textbf{#1}\par\smallskip}
\newcolumntype{Y}{>{\raggedright\arraybackslash}p{\dimexpr\linewidth-52pt\relax}}
\newenvironment{pmrows}{\noindent\renewcommand{\arraystretch}{1.25}\begin{tabular}{@{}l@{\hspace{5pt}}Y@{}}}{\end{tabular}\par}
\newsavebox{\pmsave}
\newenvironment{pmpanel}{\par\noindent\footnotesize\setlength{\fboxsep}{6pt}\begin{lrbox}{\pmsave}\begin{minipage}{\dimexpr\linewidth-2\fboxsep-2\fboxrule\relax}}{\end{minipage}\end{lrbox}\fcolorbox{pmframe}{pmback}{\usebox{\pmsave}}\par}

\setcounter{page}{1}
\maketitlesupplementary

\renewcommand{\topfraction}{0.9}
\renewcommand{\bottomfraction}{0.5}
\renewcommand{\textfraction}{0.1}
\renewcommand{\floatpagefraction}{0.85}
\renewcommand{\dbltopfraction}{0.9}
\renewcommand{\dblfloatpagefraction}{0.85}
\setcounter{topnumber}{3}
\setcounter{bottomnumber}{2}
\setcounter{totalnumber}{5}
\setcounter{dbltopnumber}{2}

\setcounter{figure}{0}
\setcounter{table}{0}
\setcounter{section}{0}
\setcounter{prompt}{0}
\renewcommand{\thesection}{S\arabic{section}}
\renewcommand{\thefigure}{S\arabic{figure}}
\renewcommand{\thetable}{S\arabic{table}}
\renewcommand{\theprompt}{S\arabic{prompt}}

In the supplementary material, we provide:
\begin{itemize}
  \item \textbf{Implementation Details} (Sec.~\ref{sec:imp}): settings, mark types, prompts, and the settings of Tab.~5.
  \item \textbf{Data} (Sec.~\ref{sec:data}): training data, region labels, Tangled Lines, and the Line Tracing set of KLAL.
  \item \textbf{Evaluation} (Sec.~\ref{sec:eval}): protocols, the re-evaluation of DeepEyes and Pixel-Reasoner, and latency.
  \item \textbf{Visualizations} (Sec.~\ref{sec:show}): the three steps on each task family, and failure cases.
\end{itemize}
Tables, figures, and sections without the prefix S refer to the main paper.

\section{Implementation Details}
\label{sec:imp}

\subsection{Model and Training}
\label{sec:train}

Tab.~\ref{tab:hyper} summarizes the implementation settings.
We fine-tune Qwen2.5-VL-7B~\cite{bai2025qwen2} using two LoRA adapters~\cite{hu2022lora}, one for the locate pass and the other for the remaining passes, while keeping the vision encoder frozen.
We add four type tokens to the vocabulary and use their embeddings as $\mathbf{e}_\tau$.
The editor is a transformer that maps patch-wise conditions to mark features in a single forward pass, with separate input channel groups for each mark type.
It is initialized from an editor pretrained to draw points and lines using only $\mathcal{L}_{\text{mark}}$.
For counting, high-resolution VQA, and line tracing, the locate pass uses $1792\times1792$ images to capture small objects and thin curves.
High-resolution VQA also uses this resolution for the read pass.

\begin{table}[t]
  \centering
  \caption{\textbf{Hyperparameters.} Shared settings are listed once.}
  \label{tab:hyper}
  \footnotesize
  \setlength{\tabcolsep}{3pt}
  \begin{tabular}{@{}lll@{}}
    \toprule
    & Qwen2.5-VL-7B & Qwen3.5-9B \\
    \midrule
    \multicolumn{3}{@{}l}{\textit{Model}} \\
    Input size & $896^2$ & $1024^2$ \\
    Locate input size$^\dagger$ & $1792^2$ & $2048^2$ \\
    LoRA & \multicolumn{2}{l}{2 adapters, rank 16, $\alpha=32$, all linear layers} \\
    Editor & \multicolumn{2}{l}{8 blocks, width 1,536, 12 heads} \\
    Locate heads & \multicolumn{2}{l}{two-layer MLP of width 1,024 per mark type} \\
    Parameters\textsuperscript{\S} & 81M / 15M / 0.39B & 87M / 17M / 0.39B \\
    \midrule
    \multicolumn{3}{@{}l}{\textit{Training}} \\
    Optimizer & \multicolumn{2}{l}{AdamW, weight decay 0.01, one cycle, 5\% warm-up} \\
    Peak learning rate & \multicolumn{2}{l}{$10^{-4}$ adapters, $10^{-3}$ heads, $3\times10^{-5}$ editor} \\
    Gradient clipping; $\beta$, $\gamma$ & \multicolumn{2}{l}{1.0 per parameter group; 5, 5} \\
    BCE positive weight & \multicolumn{2}{l}{3 for paths and boxes, 5 for lines} \\
    Binary conditions & \multicolumn{2}{l}{paths and lines at 0.5, straight-through gradient} \\
    Steps, batch size & \multicolumn{2}{l}{15,153, 16} \\
    Epochs$^\ddagger$ & \multicolumn{2}{l}{20 / 10 / 3 / 4} \\
    Answer tokens$^\ddagger$ & \multicolumn{2}{l}{at most 512 / 256 / 256 / 16} \\
    DyME & \multicolumn{2}{l}{8 answers, temperature 1.0, ramp over 2,400 steps} \\
    Time on 16 B200 GPUs & 27\,h & 53\,h \\
    \midrule
    \multicolumn{3}{@{}l}{\textit{LoRA baseline}} \\
    Steps & \multicolumn{2}{l}{15,153 supervised, then 6,000 GRPO} \\
    \bottomrule
  \end{tabular}\\[2pt]
  {\scriptsize $^\dagger$Counting, high-res.\ VQA, and line tracing. \textsuperscript{\S}Adapters / heads / editor. $^\ddagger$Spatial planning / counting / high-res.\ VQA / line tracing.}
\end{table}

All four task families are trained jointly, with each training step sampling a family in proportion to its dataset size.
For half of the samples, the read pass uses marks generated from ground-truth conditions, exposing the reader to accurate marks during training.
For counting, 75\% of the samples are presented as $2\times2$ or $3\times3$ mosaics of same-category images to increase the count range.
Following DyME~\cite{liu2025empowering}, we generate 8 answers per sample and apply GRPO~\cite{shao2024deepseekmath} if any answer is correct, or $\mathcal{L}_{\text{ans}}$ otherwise.
GRPO advantages are standardized across the 8 answers, without a KL penalty.
A spatial plan is considered correct if it reaches the goal without entering a hole.
For counting, a prediction $n$ is considered correct if its reward $\exp(-|\ln((n+1)/(y+1))|)$ is at least 0.9, where $y$ is the ground-truth count.

The LoRA baseline trains an adapter of the same rank on the reference answers of the same items, without marks or mosaics, and then runs 6,000 GRPO steps with the same rewards.
For Qwen3.5-9B~\cite{qwen2026qwen35}, images are resized to $1024\times1024$ to keep the $32\times32$ grid, and the editor is trained from scratch, first with $\mathcal{L}_{\text{mark}}$ alone for 2,000 steps.

\subsection{Mark Types}
\label{sec:types}

Tab.~\ref{tab:types} summarizes the four mark types.
The labeled marks are drawn in red on a white canvas, which the frozen vision encoder turns into $\mathbf{Z}^*$.
For boxes, the box around the marked sub-cells is enlarged to a square of at least 2.5 times its size, cropped from the full-resolution image, and encoded, and the question is repeated after the crop.

\begin{table*}[t]
  \centering
\caption{\textbf{Mark types.} At $896\times896$, each sub-cell covers $7\times7$ pixels. At $1792\times1792$, the locate pass predicts $2\times2$ sub-cells per visual token, which are regrouped into $4\times4$ sub-cells per patch.}
  \label{tab:types}
  \footnotesize
  \setlength{\tabcolsep}{5pt}
  \begin{tabular}{@{}lllllll@{}}
    \toprule
    Type & Task & Locate input & Condition per sub-cell & Label & Drawn for $\mathbf{Z}^*$ & Loss $\ell_\tau$ \\
    \midrule
    Points, \texttt{<PTS>} & counting & $1792^2$ & expected objects & annotated points & dot, radius 2 px & Poisson NLL \\
    Paths, \texttt{<PATH>} & spatial planning & $896^2$ & route coverage & squares of all shortest routes & square & BCE \\
    Boxes, \texttt{<BOX>} & high-res.\ VQA & $1792^2$ & region coverage & region boxes (Sec.~\ref{sec:regions}) & none, read as a crop & BCE \\
    Lines, \texttt{<LINE>} & line tracing & $1792^2$ & curve coverage & curve from the asked box & stroke, 12 px wide & BCE \\
    \bottomrule
  \end{tabular}
\end{table*}

\subsection{Prompts}
\label{sec:prompts}

Prompt~\ref{prm:tasks} shows the questions and answers.
The answer of \model{} starts with the type token, which is followed by the marks.
For spatial planning, \model{} lists the squares that it reads from the marks before the plan, and the LoRA baseline outputs the plan directly, as in the direct fine-tuning of Mirage~\cite{yang2025mirage}.
Zero-shot MLLMs receive the same questions.

\begin{prompt}[t]
  \begin{pmpanel}
  \pmtask{Spatial planning (VSP)}
  \begin{pmrows}
    \pmuser & As a professional maze solver, your task is to analyze a grid-based map and devise an action plan that enables a player to reach the goal from the starting point without falling into any holes, using the fewest possible moves. \pmcut{game setup and moving rules of Mirage} Here is the map: (the image above) Please provide your action plan. The final answer MUST BE put in \pmbox{}. \\
    \pmours & \pmtok{<PATH>} \pmvis{marks} A safe path from the player to the goal visits these squares in order: row 1, column 3; row 2, column 3; row 2, column 2; row 2, column 1; row 1, column 1. \pmbox{D, L, L, U} \\
    \pmbase & \pmbox{D, L, L, U} \\
  \end{pmrows}
  \pmtask{Dense counting (FSC-147)}
  \begin{pmrows}
    \pmuser & How many cashew nuts are in the image? Answer with a single number only. \\
    \pmours & \pmtok{<PTS>} \pmvis{marks} 157 \\
    \pmbase & 157 \\
  \end{pmrows}
  \pmtask{High-resolution VQA (V*, HR-Bench)}
  \begin{pmrows}
    \pmuser & What is the color of the helmet? (A) green (B) blue (C) red (D) white Answer with the option's letter from the given choices directly. \\
    \pmours & \pmtok{<BOX>} \pmvis{crop} \pmcut{the question again} A \\
    \pmbase & A \\
  \end{pmrows}
  \pmtask{Line tracing (Tangled Lines)}
  \begin{pmrows}
    \pmuser & Which node is connected to node J? Answer with the letter. \\
    \pmours & \pmtok{<LINE>} \pmvis{marks} \pmbox{I} \\
    \pmbase & \pmbox{I} \\
  \end{pmrows}
  \end{pmpanel}
  \caption{\textbf{Questions and answers.} The image precedes the question. \pmvis{marks} and \pmvis{crop} stand for the visual tokens after the type token, \pmcut{\dots} for omitted text, and LoRA for the LoRA baseline. VQA training questions without options keep the open-ended form of DeepEyes.}
  \label{prm:tasks}
\end{prompt}

\subsection{Settings of Tab.~5}
\label{sec:ablation}

The upper part of Tab.~5 reads the model of Tab.~3 in different ways at test time.
None gives the editor empty conditions, and for boxes a blank crop.
Mirrored flips the predicted conditions left to right before the editor draws them, and for boxes zooms into the region at the mirror position of the located one.
Ground truth draws the marks from the annotated conditions, and for boxes crops the annotated region.
The lower part retrains the model with the settings of Tab.~\ref{tab:hyper}, changed as follows.
Without marking, the locate pass runs on the reader's adapter, so that $\mathcal{L}_{\text{loc}}$ reaches the weights that answer, and the reader answers from the image and the question alone, with no type token, marks, or zoom.
Without $\mathcal{L}_{\text{mark}}$, the editor starts from random weights and learns only from the answers.

\section{Data}
\label{sec:data}

\subsection{Training Data}
\label{sec:traindata}

Each task family trains only on the data of the compared state of the art (Tab.~\ref{tab:data}).
Counting also uses the point annotations of FSC-147~\cite{ranjan2021fsc} for $\mathcal{L}_{\text{loc}}$ and $\mathcal{L}_{\text{mark}}$.
We do not compare with Mini-o3~\cite{lai2025minio3}, which adds its own VisualProbe questions and multi-turn trajectories to those of DeepEyes.

\begin{table}[t]
  \centering
  \caption{\textbf{Training data.} \model{} and the LoRA baseline train once on the union of the four sets. $^\ddagger$Re-implemented on Tangled Lines.}
  \label{tab:data}
  \footnotesize
  \setlength{\tabcolsep}{3pt}
  \begin{tabular}{@{}p{0.30\linewidth}p{0.65\linewidth}@{}}
    \toprule
    Method & Training data \\
    \midrule
    \multicolumn{2}{@{}l}{\textit{Spatial planning}} \\
    Mirage~\cite{yang2025mirage} & 1,000 VSP maps with helper images \\
    CoLVR~\cite{ding2026colvr} & the 1,000 maps of Mirage and 500 maps of its RL stage \\
    \rowcolor{gray!15}
    \model{} & the maps of CoLVR, 1,497 after removing duplicates \\
    \midrule
    \multicolumn{2}{@{}l}{\textit{Dense counting}} \\
    WS-COC~\cite{zhang2026wscoc} & 3,659 FSC-147 training images with counts \\
    \rowcolor{gray!15}
    \model{} & the same images with counts and points \\
    \midrule
    \multicolumn{2}{@{}l}{\textit{High-resolution VQA}} \\
    DeepEyes~\cite{zheng2025deepeyes} & 47K fine-grained, chart, and reasoning questions \\
    Pixel-Reasoner~\cite{su2025pixelreasoner} & its own instruction-tuning and RL data \\
    \rowcolor{gray!15}
    \model{} & 18,644 fine-grained questions of DeepEyes \\
    \midrule
    \multicolumn{2}{@{}l}{\textit{Line tracing}} \\
    KLAL$^\ddagger$ & 30,000 Tangled Lines images \\
    \rowcolor{gray!15}
    \model{} & the same images \\
    \bottomrule
  \end{tabular}
\end{table}

\subsection{Region Labels for High-Resolution VQA}
\label{sec:regions}

DeepEyes~\cite{zheng2025deepeyes} releases its training questions without the regions that answer them.
We let the base MLLM ground these regions (Prompt~\ref{prm:ground}) and keep the 18,644 questions whose region covers less than a quarter of the image.
The first pass often boxes only one of two related objects, so a second pass grounds every object that the question mentions.
The resulting zoom contains the target for 88.0\% of the V* questions, and the trained locate step reaches 91.1\%.
Half of the questions with a short answer are posed as multiple choice, the format of the benchmarks.

\begin{prompt}[t]
  \begin{pmpanel}
  \begin{pmrows}
    \pmchip{pmuserbg}{pmuser}{Pass 1} & Question: \{question\} Locate the object or objects in the image that this question is about. Output their bounding boxes in JSON format: [\{"bbox\_2d": [x1, y1, x2, y2], "label": "name"\}]. \\
    \pmchip{pmuserbg}{pmuser}{Pass 2} & Question: \{question\} Locate every object that this question mentions, including any object it uses as a reference, each one separately. Output their bounding boxes in JSON format: [\{"bbox\_2d": [x1, y1, x2, y2], "label": "name"\}]. \\
  \end{pmrows}
  \end{pmpanel}
  \caption{\textbf{Grounding prompts} for the regions of the VQA training questions. Pass 1 selects the questions, and pass 2 gives the labels of the locate step.}
  \label{prm:ground}
\end{prompt}

\subsection{Tangled Lines}
\label{sec:tangle}

Each $768\times768$ image holds 3 to 5 smooth curves that join pairs of lettered boxes on its border (Fig.~\ref{fig:data}a).
The question asks which box is joined to a given one, so chance is $1/(2k-1)$ for $k$ curves, 15.0\% over the test set.
We generate 30,000 training and 600 test images from different random seeds, so every test image has a new layout.
The labeled mark is the curve that leaves the asked box.
The test set and the generator code are included with this supplementary material.

\subsection{The Line Tracing Set of KLAL}
\label{sec:klal}

\begin{figure*}[t]
  \centering
  \includegraphics[width=0.9\textwidth]{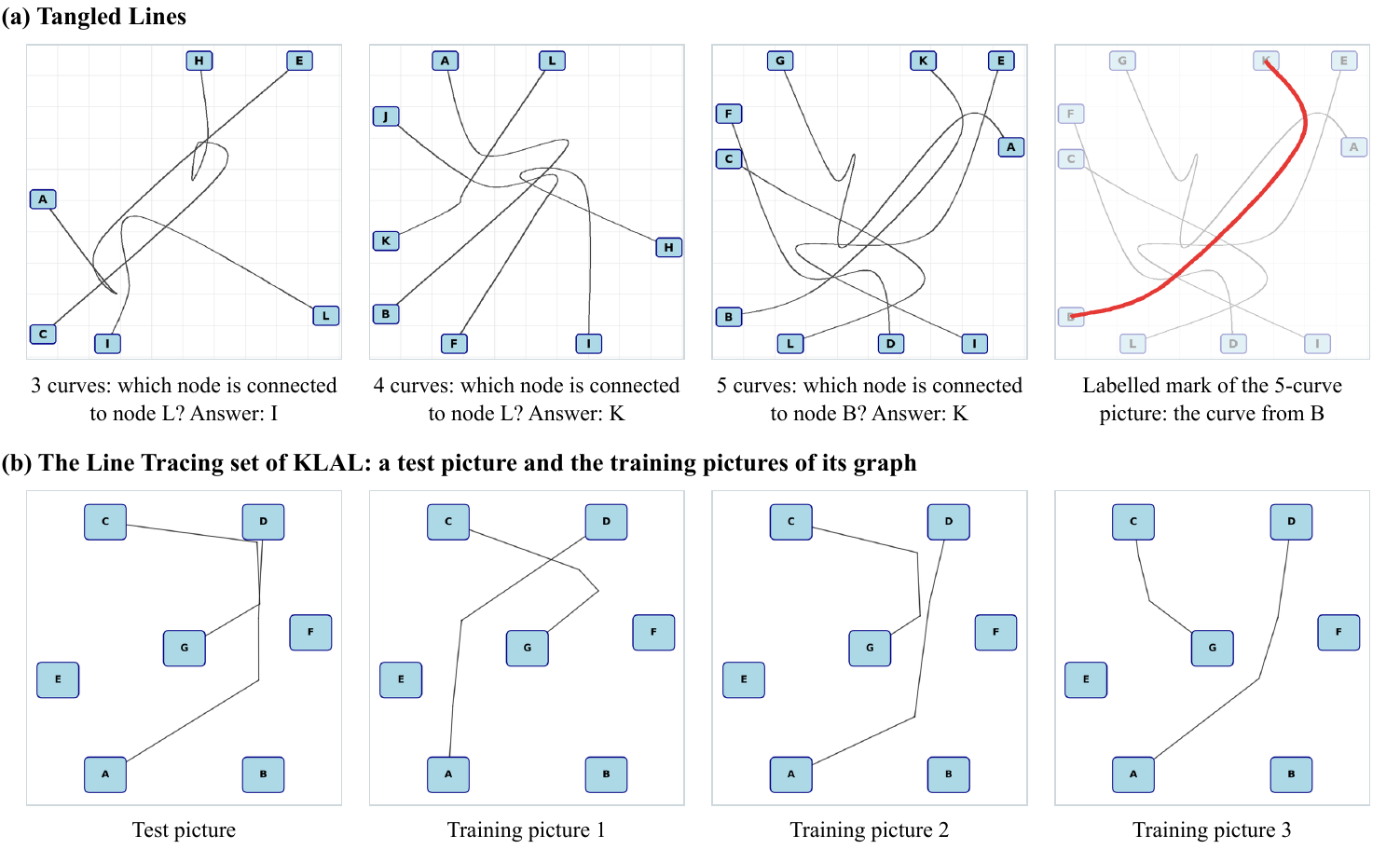}
  \caption{\textbf{Line tracing data.} (a) Tangled Lines with 3, 4, and 5 curves, and the labeled mark of the last image. (b) A test image of the Line Tracing set of KLAL and the three training images of its graph, which differ only in the bends of the curves.}
  \label{fig:data}
\end{figure*}

The Line Tracing set of KLAL~\cite{esmaeilkhani2026klal} asks whether two boxes are joined by one of 2 or 3 polygonal curves.
Of its 210 test images, 206 redraw a training graph with the same boxes and the same joined pairs (Fig.~\ref{fig:data}b), and 707 of its 1,048 test questions appear in the training set with the same answer.
Trained on this set alone, the LoRA baseline reaches 96.9\% (Tab.~\ref{tab:klal}).
After a new split by graph, a LoRA fine-tune still reaches 95.4\%, which leaves no room to compare methods, so we built Tangled Lines.

\begin{table}[t]
  \centering
  \caption{\textbf{The Line Tracing set of KLAL.} Accuracy (\%) on its yes-or-no test questions, chance 50\%. Our models are trained on this set alone.}
  \label{tab:klal}
  \footnotesize
  \begin{tabular}{@{}lc@{}}
    \toprule
    Method & Accuracy \\
    \midrule
    \multicolumn{2}{@{}l}{\textit{Reported by KLAL~\cite{esmaeilkhani2026klal}}} \\
    GPT-4o / Gemini-2.0 Flash & 55.3 / 59.3 \\
    Qwen2.5-VL-7B, zero-shot / fine-tuned / with KLAL & 49.6 / 53.8 / 62.2 \\
    \midrule
    \multicolumn{2}{@{}l}{\textit{Ours, Qwen2.5-VL-7B}} \\
    LoRA baseline & 96.9 \\
    \model{} / without marks & 96.3 / 96.2 \\
    LoRA fine-tune, split by graph & 95.4 \\
    \bottomrule
  \end{tabular}
\end{table}

\section{Evaluation}
\label{sec:eval}

\subsection{Protocol}
\label{sec:protocol}

Tab.~\ref{tab:protocol} lists the test sets, decoding, and answer parsing.
V* and HR-Bench follow Mini-o3~\cite{lai2025minio3} and average 32 and 8 answers sampled at temperature 1.0.
Zero-shot MLLMs use the same questions and parsing.
Qwen3-VL-8B-Thinking~\cite{bai2025qwen3vl} samples with its released configuration and thinks for up to 40,960 tokens.

\begin{table}[t]
  \centering
  \caption{\textbf{Evaluation protocol.} FSC-147 has 1,286 validation and 1,190 test images. $^\dagger$An answer without a number, or with a number above $10^7$, counts as 0.}
  \label{tab:protocol}
  \footnotesize
  \setlength{\tabcolsep}{4pt}
  \begin{tabular}{@{}lrll@{}}
    \toprule
    Benchmark & Items & Decoding & Answer \\
    \midrule
    VSP & 400 & greedy & boxed plan, simulated \\
    FSC-147 & 1,286 / 1,190 & greedy & first number$^\dagger$ \\
    V* & 191 & Avg@32 & option letter \\
    HR-Bench 4K / 8K & 800 / 800 & Avg@8 & option letter \\
    Tangled Lines & 600 & greedy & boxed letter \\
    \bottomrule
  \end{tabular}
\end{table}

\subsection{Re-evaluation of DeepEyes and Pixel-Reasoner}
\label{sec:reeval}

We re-evaluate DeepEyes~\cite{zheng2025deepeyes} and Pixel-Reasoner~\cite{su2025pixelreasoner} using their released checkpoints and zooming pipelines (Tab.~\ref{tab:reeval}).
We find that the DeepEyes evaluation script places the correct option first, yielding 86.9\% on V*, compared with 85.3\% when the options are shuffled.
On the released question files, our DeepEyes results closely match those independently reported in~\cite{lai2025minio3}, with 83.0\% on V* and 73.3\%/69.7\% on HR-Bench 4K/8K.
Our Pixel-Reasoner evaluation also yields comparable results, reaching 85.9\% on V* against the reported 84.3\%.
We use the re-evaluated results on the released questions for the main comparison.

\begin{table}[t]
  \centering
  \caption{\textbf{Re-evaluation of tool-based baselines.}
  Accuracy (\%) in our environment.
  Bold indicates results used in Tab.~3c.}
  \label{tab:reeval}
  \footnotesize
  \setlength{\tabcolsep}{3pt}
  \begin{tabular}{@{}lcccl@{}}
    \toprule
    Evaluation & V* & HR 4K & HR 8K & Prior evaluation \\
    \midrule
    \multicolumn{5}{@{}l}{\textit{DeepEyes}} \\
    Original script & 86.9 & -- & -- &
      89.0~\cite{deepeyes2025pr66} \\
    Shuffled options & 85.3 & -- & -- &
      84.3--85.9~\cite{deepeyes2025pr66} \\
    Released questions & \textbf{83.0} &
      \textbf{73.3} & \textbf{69.7} &
      83.3 / 73.2 / 69.5~\cite{lai2025minio3} \\
    \midrule
    \multicolumn{5}{@{}l}{\textit{Pixel-Reasoner}} \\
    Released questions & \textbf{85.9} &
      \textbf{72.4} & \textbf{64.9} &
      84.3~\cite{su2025pixelreasoner} \\
    Alternative decoding & 80.8 &
      71.0 & 65.5 & -- \\
    \bottomrule
  \end{tabular}
\end{table}

\subsection{Latency}
\label{sec:latency}

Tab.~\ref{tab:latency} gives the numbers behind Fig.~3, timed from the decoded image to the answer.
Locating and marking take 0.1 to 0.7\,s per question.
\model{} generates 2 to 6 tokens per answer, except for the 58 tokens on VSP that list the route squares, while Qwen3-VL-8B-Thinking generates 196 to 7,471.

\begin{table}[t]
  \centering
  \caption{\textbf{Latency.} Seconds per question on one B200 GPU at a batch size of one. Encode includes emitting the type token; for boxes, mark is the crop and its encoding.}
  \label{tab:latency}
  \footnotesize
  \setlength{\tabcolsep}{3.2pt}
  \begin{tabular}{@{}lrrrrr@{}}
    \toprule
    Method & V* & HR-8K & FSC-147 & VSP & Tangled \\
    \midrule
    Qwen2.5-VL-7B & 0.46 & 2.08 & 0.09 & 6.93 & 0.10 \\
    Qwen3-VL-8B-Thinking & 5.07 & 21.66 & 72.65 & 189.11 & 27.20 \\
    DeepEyes & 3.83 & 8.93 & -- & -- & -- \\
    Pixel-Reasoner & 4.86 & 5.00 & -- & -- & -- \\
    WS-COC & -- & -- & 0.17 & -- & -- \\
    LoRA baseline & 0.51 & 0.72 & 0.19 & 0.39 & 0.28 \\
    \rowcolor{gray!15}
    \model{} & 1.08 & 1.56 & 0.77 & 2.11 & 0.88 \\
    \midrule
    \multicolumn{6}{@{}l}{\textit{\model{} by step}} \\
    Encode & 0.49 & 0.70 & 0.13 & 0.13 & 0.14 \\
    Locate & 0.27 & 0.26 & 0.49 & 0.07 & 0.51 \\
    Mark & 0.13 & 0.41 & 0.01 & 0.01 & 0.01 \\
    Read & 0.20 & 0.19 & 0.14 & 1.90 & 0.23 \\
    \bottomrule
  \end{tabular}
\end{table}

\section{Visualizations}
\label{sec:show}

\begin{figure*}[p]
  \centering
  \includegraphics[width=\textwidth]{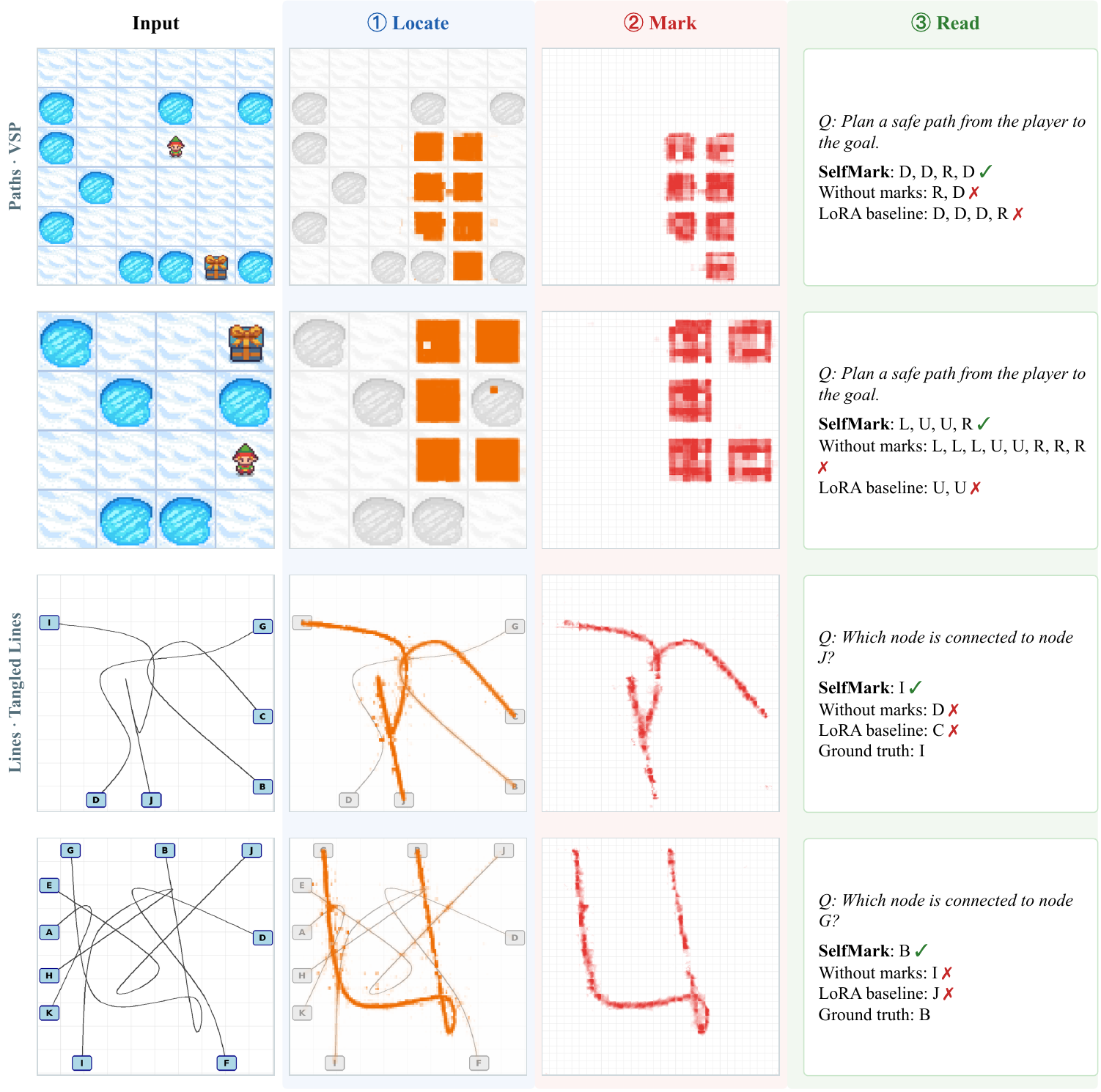}
  \caption{\textbf{Spatial planning and line tracing.} (1) Locate: the predicted coverage in orange. (2) Mark: the mark features after the type token, decoded into pixels by a linear probe for display. (3) Read: answers with the marks, without them, and of the LoRA baseline; the VSP question is abbreviated.}
  \label{fig:paths}
\end{figure*}

\begin{figure*}[p]
  \centering
  \includegraphics[width=\textwidth]{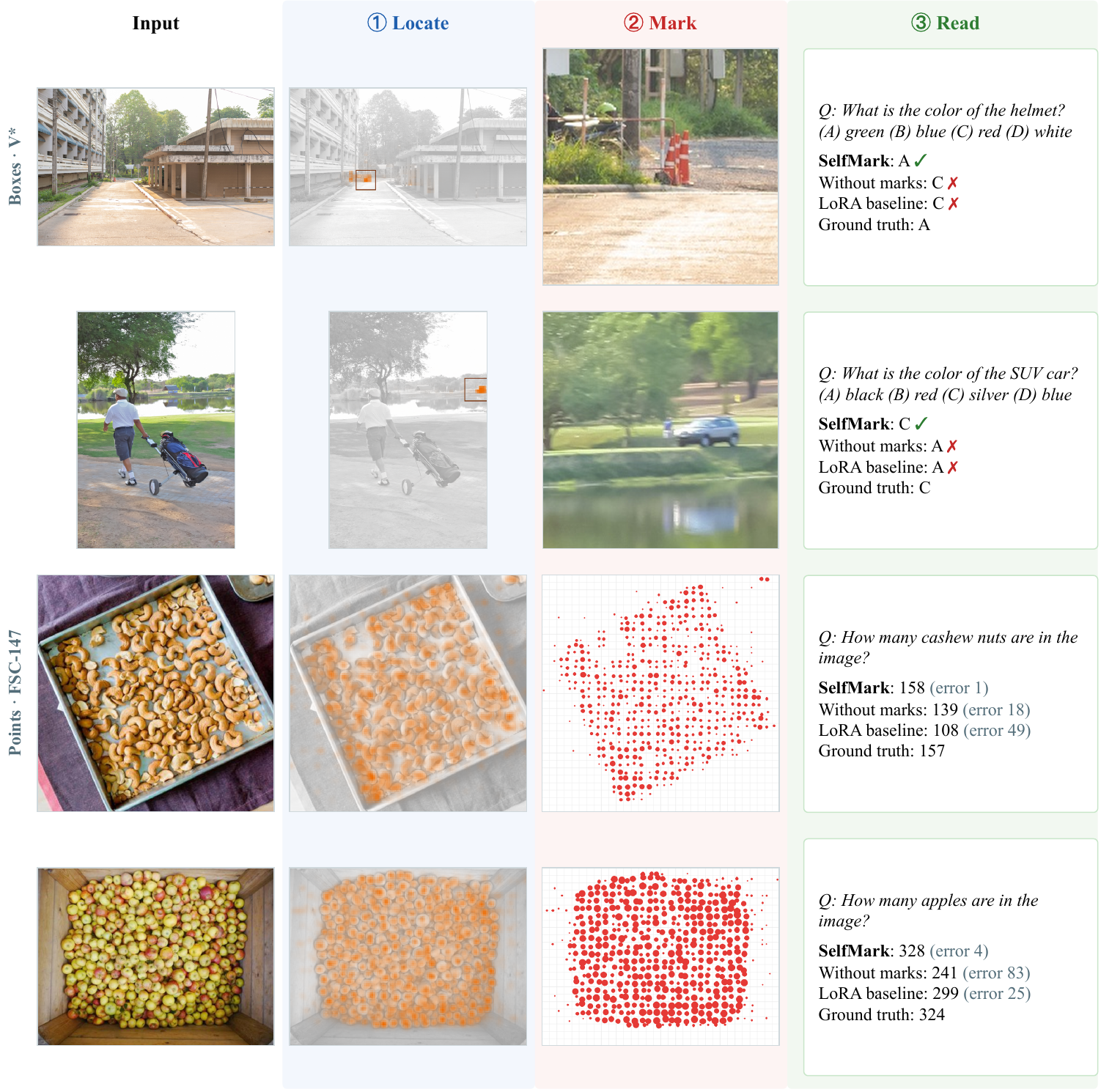}
  \caption{\textbf{High-resolution VQA and counting.} (1) Locate: the predicted coverage on V*, with the zoom outlined, and the expected counts on FSC-147. (2) Mark: the crop on V*, and on FSC-147 the expected count of each patch as a dot of proportional area. (3) Read: as in Fig.~\ref{fig:paths}, with the error of each count.}
  \label{fig:boxes}
\end{figure*}

\begin{figure*}[p]
  \centering
  \includegraphics[width=\textwidth]{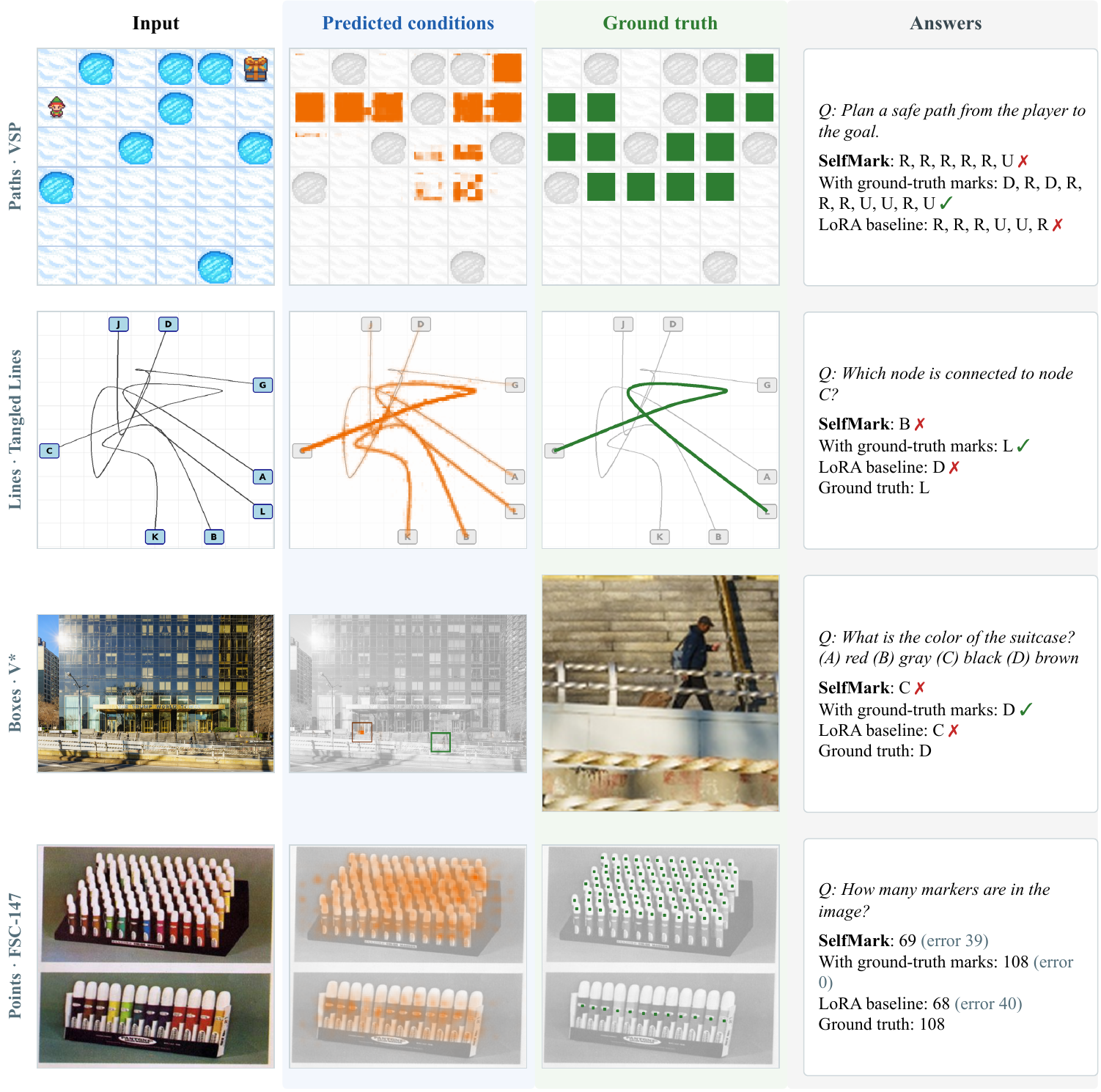}
  \caption{\textbf{Failure cases.} The predicted conditions in orange against the ground truth in green; on V*, both zooms are outlined, and the ground-truth crop is shown. Each question is answered correctly with the marks of the ground-truth conditions.}
  \label{fig:fail}
\end{figure*}

Figs.~\ref{fig:paths} and~\ref{fig:boxes} visualize the three steps on test examples where \model{} outperforms the LoRA baseline, while Fig.~\ref{fig:fail} presents failure cases. All examples use seed 0.
To visualize the mark features, we fit a linear probe using ridge regression ($\lambda=10$) on 96 training images. The probe maps standardized token features to the $28\times28$ pixels of each patch, supervised by the corresponding labeled marks.
It is used only for visualization; the MLLM reads the mark features directly.
For counting, the conditions are fractional, so we visualize the expected count in each patch as a dot with proportional area.

In the successful examples, the marks highlight the route, trace the target curve through crossings, identify the queried object, or locate the objects to count.
Removing these marks causes the same model to answer incorrectly.
The four failure cases illustrate different localization errors: the route enters a hole, the traced line switches curves at a crossing, the zoom targets the wrong object, and densely packed markers are undercounted.
All four failures originate from the locate step, as the model answers each question correctly when given marks generated from ground-truth conditions.

\clearpage
{
  \small
  \bibliographystyle{ieeenat_fullname}
  \bibliography{main}
}

\end{document}